\documentclass[sigconf]{acmart}
\usepackage{subcaption}
\usepackage{enumitem}
\usepackage{calc}
\usepackage{amsthm}
\AtBeginDocument{%
  \providecommand\BibTeX{{%
    \normalfont B\kern-0.5em{\scshape i\kern-0.25em b}\kern-0.8em\TeX}}}
\newtheorem{theorem}{Theorem}

\renewcommand\footnotetextcopyrightpermission[1]{} 
\begin{document}

\title[Multi-Level Time-series Disentanglement]{Interpretable Time-series Representation Learning\\ With Multi-Level Disentanglement}

\newcommand{\modele}{\textbf{D}isentangle \textbf{T}ime-\textbf{S}eries~}
\newcommand{\model}{\texttt{DTS}~}
\newcommand{\modelnospace}{\texttt{DTS}}
\newcommand{\zdh}[1]{\textcolor{blue}{[***#1***]}}
\newcommand{\yn}[1]{\textcolor{red}{[***#1***]}}

\newcommand{\jni}[2]{\textcolor{orange}{#1 (jni: #2)}}
\newcommand{\rev}[1]{\textcolor{orange}{#1}}

\newcommand{\nop}[1]{}

\vspace{-30pt}

\begin{abstract}
Time-series representation learning is a fundamental task for time-series analysis. While significant progress has been made to achieve accurate representations for downstream applications, the learned representations often lack interpretability and do not expose semantic meanings. Different from previous efforts on the entangled feature space, we aim to extract the semantic-rich temporal correlations in the latent interpretable factorized representation of the data.
Motivated by the success of disentangled representation learning in computer vision, 
we study the possibility of learning semantic-rich time-series representations, which remains unexplored due to three main challenges: 1) sequential data structure introduces complex temporal correlations and makes the latent representations hard to interpret, 2) sequential models suffer from KL vanishing problem, and 3) interpretable semantic concepts for time-series often rely on multiple factors instead of individuals. To bridge the gap, we propose ~\modele (\modelnospace), a novel disentanglement enhancement framework for sequential data. Specifically, to generate hierarchical semantic concepts as the interpretable and disentangled representation of time-series, ~\model introduces multi-level disentanglement strategies by covering both individual latent factors and group semantic segments. We further propose an evidence lower bound (ELBO) decomposition approach to balance the preference between correct inference and fitting data distribution problems. We theoretically show how to alleviate the KL vanishing problem: \model introduces a mutual information maximization term, while preserving a heavier penalty on the total correlation and the dimension-wise KL to keep the disentanglement property. Experimental results on various real-world datasets demonstrate that the representations learned by ~\model achieve superior performance in downstream applications, with high interpretability of semantic concepts. 
\end{abstract}

\author{Yuening Li$^1$, Zhengzhang Chen$^2$, Daochen Zha$^1$,
Mengnan Du$^1$,
Denghui Zhang$^3$, Haifeng Chen$^2$, Xia Hu$^1$}
\affiliation{
\institution{$^1$Department of Computer Science and Engineering, Texas A\&M University} 
\institution{$^2$NEC Labs America}
\institution{$^3$Rutgers University}
}
\email{liyuening@tamu.edu}

 
\settopmatter{printacmref=false}

\maketitle

\section{Introduction}


\begin{figure}[t]
      \begin{subfigure}[b]{0.4\textwidth}
 \includegraphics[width=\linewidth]{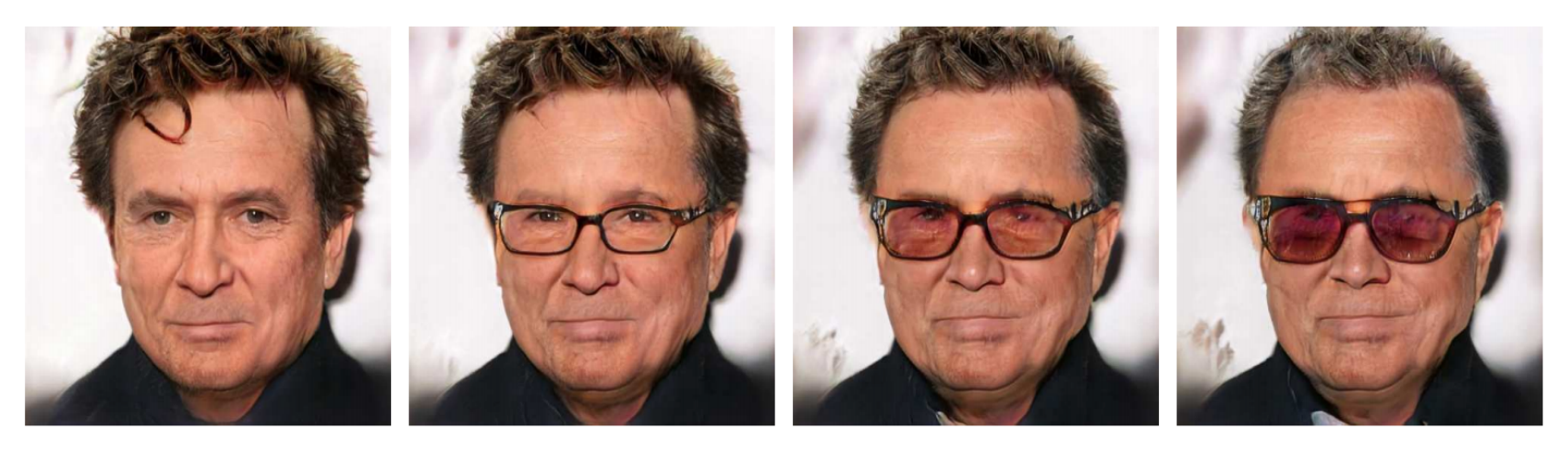}
 \caption{Semantic factors for images, where a semantic factor controls the eye glasses of a human facial image~\cite{shen2020interfacegan}. }

        \end{subfigure}%
        \newline
  \begin{subfigure}[b]{0.4\textwidth} \includegraphics[width=\linewidth]{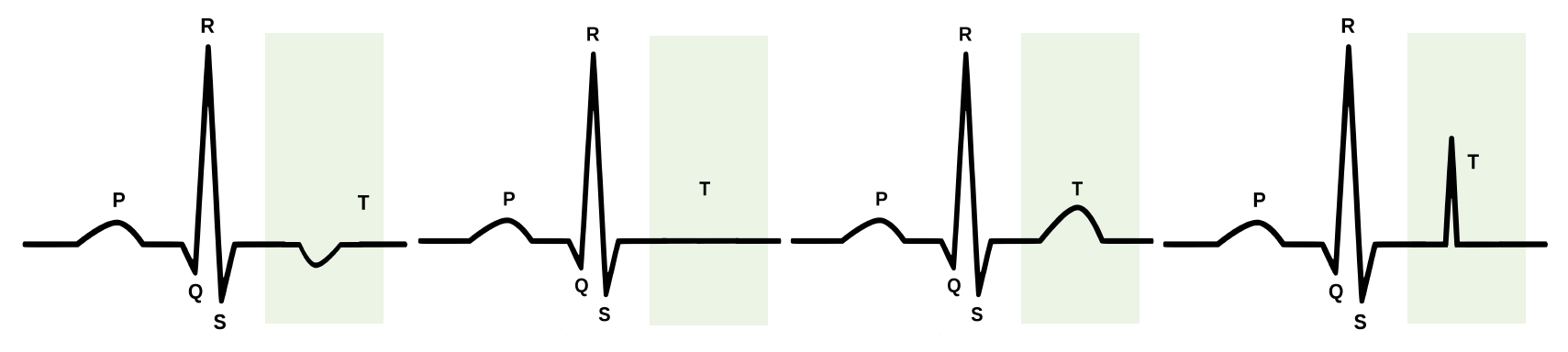}
 \caption{Semantic factors of time-series, where a semantic factor controls the sequential trend of a time-series. }
        \end{subfigure}%
        \vspace{-5pt}
        \caption{Two traversal plot examples of disentanglement. }\label{fig:traversalintro}
\end{figure}

Time-series data are ubiquitous in various domains, 
such as self-driving vehicles, financial transactions, and computer-aided medical diagnosis~\cite{blackburn1960electrocardiogram,lai2020tods,zha2020meta}.
As a fundamental task of time-series analysis, unsupervised time-series representation learning aims to extract low-dimensional representations from complex raw time-series without human supervision. These representations can then be used to benefit many downstream tasks, such as classification~\cite{liu2019single,huang2019graph}, clustering~\cite{du2019towards,lai2020dual,zhou2020towards}, and anomaly detection~\cite{li2019specae,li2020pyodds,li2019deep,li2020autood}. 
Recently, deep generative models have shown great representation ability in modeling complex underlying distributions of time-series data. The most representative examples include LSTM-VAE and its different variants~\cite{semeniuta2017hybrid,fortuin2018som,fortuin2020gp}.


While recent techniques have been proposed to learn effective representations of time-series for downstream applications, the learned representations often lack the interpretability to expose tangible semantic meanings~\cite{yang2019xdeep}. 
In many cases, especially in high-stake domains, an interpretable representation is critical for diagnosis or decision making. 
For example, learning interpretable and semantic-rich representations is useful for decomposing the ECG into cardiac cycles with recognizable phases as independent factors. Furthermore, extracting and analyzing common sequential patterns ($i.e.$, normal sinus rhythms) from massive ECG records may assist clinicians to better understand the irregular symptoms (like sinus tachycardia, sinus bradycardia, and sinus arrhythmia). In contrast, diagnostic processes without transparency or accurate explanations may lead to suboptimal or even risky  treatments. 




To enable semantically meaningful representations, researchers in computer vision have been devoted to learning disentangled representations, which decompose the representations into subspaces and encode them as separate dimensions~\cite{guo2020interpretable}. A disentangled representation can be defined as one where single latent units are sensitive to changes in a single latent factor while being relatively invariant to changes in other factors; that is, different dimensions in the latent space are probabilistically independent. Fig.~\ref{fig:traversalintro} (a) shows an example, 
where a semantic factor controls the eyeglasses of a human facial image. Learning factors of variations in the images enables the emergence of semantic meanings in the underlying distribution~\cite{shen2020interfacegan}. Motivated by the success of disentanglement in the image domain, in this work, we explore disentangled representations for time-series data. Fig.~\ref{fig:traversalintro} (b) shows an example of how the learned semantic factor can control the shape of ECG time-series. Medically, inverted, biphasic, or flattened T wave could also provide insights into the abnormalities of the ventricular repolarization or secondary to abnormalities in ventricular depolarisation. In addition, the QT interval as a group of individual patterns from the beginning of the Q wave to the end of the T wave, could represent the physiologic reactions for the ventricles of the heart to depolarize and repolarize. Thus, there exists a vital need for the methods that can enhance the interpretability of time-series representations from the perspectives of both single factor disentanglement and group-level factor disentanglement, to induce multi-level semantics. 



However, time-series presents great challenges for learning disentangled representations. Firstly, \textit{complex data structure with temporal correlations makes latent codes hard to interpret}. Time-series data usually contain temporal correlations, which cannot be directly captured and interpreted by traditional image-focused disentangle  methods~\cite{shen2020interfacegan,shao2020controlvae,chen2018isolating}. While traditional sequential models, like LSTM or LSTM-VAE, could be used to model the temporal correlations, they neither provide interpretable predictions, as is often criticized, nor have an disentangle mechanism. Secondly, \textit{sequential model may suffer from KL vanishing problem}. Recent work~\cite{wang2019riemannian} shows that: since the LSTM generative model often has strong expressiveness, the reconstruction term in the objective will dominate the KL divergence term.  
In this case, the model would generate time-series without making effective use of the latent codes. Besides, the latent variables $Z$ will become independent of the observations $\mathbf{x}$ when the KL divergence term collapses to zero. It's referred to as the \textit{information preference} problem~\cite{chen2016variational}. Thirdly, \textit{interpretable semantic concepts often rely on multiple factors instead of individuals}. A human-understandable sequential pattern, called semantic components, is usually correlated with multiple factors. It is hard to interpret time-series with a single latent factor and its variations.


To address these challenges, we propose ~\modele (\modelnospace) for learning semantically interpretable time-series representations. To the best of our knowledge, \model is the first attempt to incorporate disentanglement strategies for time-series. 
It is one of the first to interpret time-series representations with state-of-the-art deep models. 
In particular, we design a multi-level time-series disentanglement strategy that accounts for both individual latent factor and group-level semantic segments, to generate hierarchical semantic concepts as the interpretable and disentangled representations of time-series. For individual latent factor disentanglement, \model introduces a mutual information maximization term (MIM) to encourage high mutual information between the latent codes and the original time-series. By leveraging the MIM, \model preserves the disentanglement property via evidence lower bound (ELBO) decomposition, while alleviates the KL vanishing problem. We further theoretically prove that the decomposition process can balance the preference between correct inference and fitting data distribution. For group-level semantic segment disentanglement, \model learns decomposed semantic segments that contain batches of independent latent variables via applying gradient reversal layers on irrelevant tasks. The learned group segment disentanglement can benefit many downstream tasks such as domain adaptation. 
Extensive experiments on real-world datasets
demonstrate that ~\model could provide more meaningful disentangled representations of time-series, and is quantitatively effective for downstream tasks.

The contributions of this work are summarized as follows:
\begin{itemize}[leftmargin=*]
\vspace{-5pt}
    \item We introduce a novel and challenging problem ($i.e.$, disentangle time-series) and propose \modelnospace, to incorporate disentanglement strategies for time-series representation learning task.
    \item We devise multi-level time-series disentanglement strategies, covering both individual latent factor and group-level semantic segments, to generate hierarchical semantic concepts as the interpretable and disentangled representation of time-series. 
     \item We propose an ELBO decomposition strategy to balance the preference between correct inference and fitting data distribution problem. We theoretically show how to alleviate the KL vanishing problem via introducing the mutual information term. Meanwhile, we preserve a heavier penalty on the total correlation and the dimension-wise KL to keep the disentangle property.
    \item We conduct extensive experiments on the  time-series representation and the domain-adaptation tasks, and provide insights on interpreting the latent representations with semantic meanings.
\end{itemize}













\vspace{-5pt}
\section{Methodology}
\begin{figure}[t]
    \centering
    \includegraphics[width=0.87\linewidth]{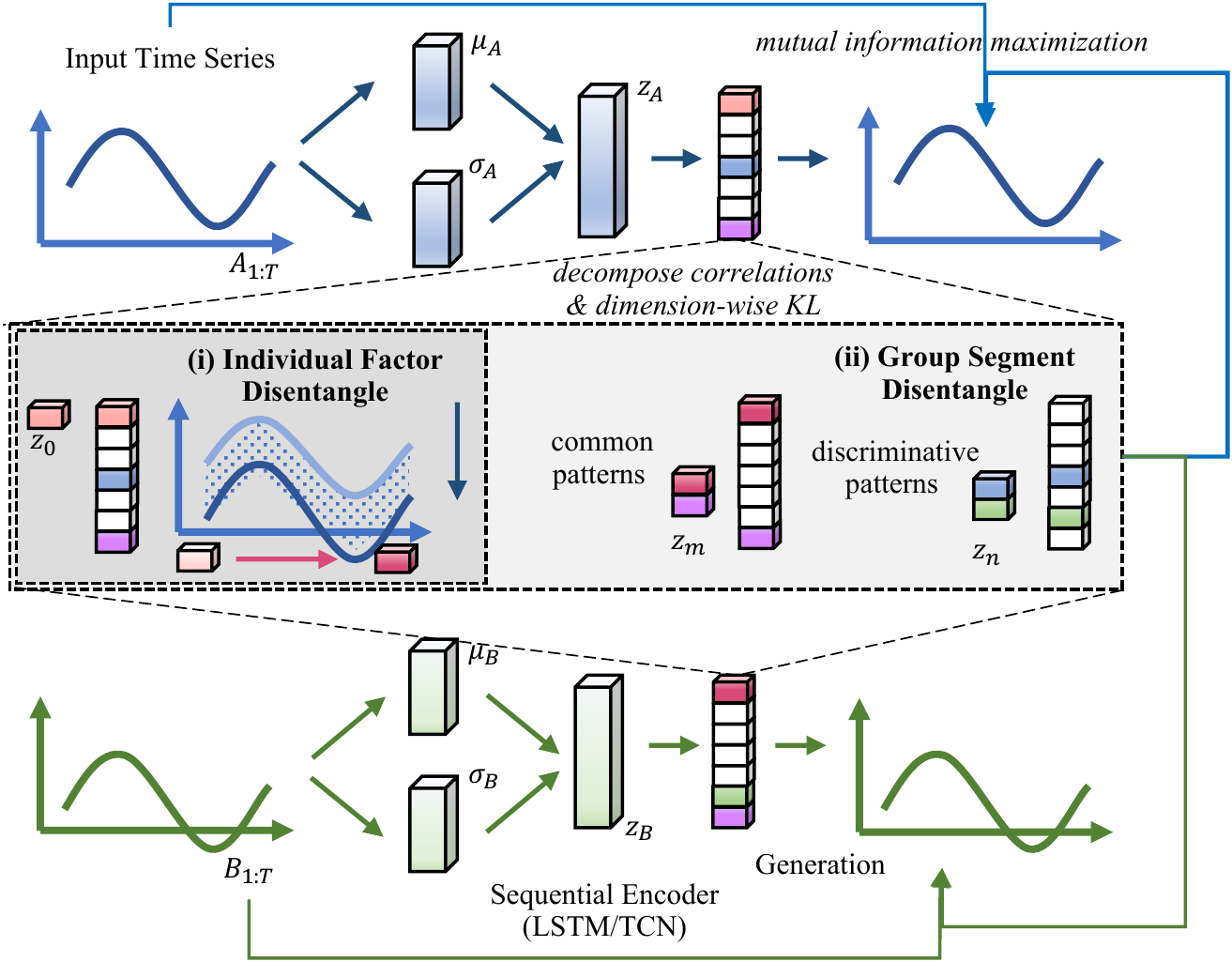}
    \caption{An overview of ~\modelnospace. It consists of two components: (i) individual factor disentangle: to learn semantic factors like $z_{0}$ (shown as pink) to control the sequential pattern of the time-series, \textit{e.g.}, the time-series moves down when adjusting $z_{0}$; and (ii) group segment disentangle: to learn more complex semantic patterns $\mathbf{z}_{m}, \mathbf{z}_{n}$ (illustrated as the common and discriminative patterns). A, B denote two time-series.}
    \label{fig:traversalintro2}
    \vspace{-10pt}
\end{figure}

In this section, we propose \model (see Fig.~\ref{fig:traversalintro2}), a multi-level disentanglement approach to enhance time-series representation learning. The key idea of ~\model is that the latent space should involve multiple independent factors as semantic concepts rather than conjugated representation. We first introduce the 
disentanglement problem (Section~2.1) and the challenges (Section~2.2). Then, we introduce an individual disentanglement approach for learning 
interpretable 
representations (Section~2.3). Finally, we introduce how to learn disentangled group segments as semantic components (Section~2.4).  

\vspace{-9pt}
\subsection{Problem Statement}
In this section, we describe the notations 
and formulate the disentanglement problem. 
To facilitate understanding, we focus on univariate time-series in the definitions; one can easily extend them to multivariate time-series. 


\noindent\textbf{Notations:} We use lowercase alphabet $x \in \mathbb{R}$ to denote a scalar value in a time-series, $z \in \mathbb{R}$ to denote a scalar latent variable, lowercase boldface letter $\mathbf{x} = [x_1, x_2, ..., x_T] \in \mathbb{R}^T$ to denote a time-series of length $T$, lowercase boldface letter $\mathbf{z} = [z_1, z_2, ..., z_N] \in \mathbb{R}^N$ to denote a segment of latent variables with size $N$, uppercase alphabet $Z$ to denote a set of latent variables, which either consists of segments, \textit{e.g.}, $Z=\{\mathbf{z}_1, \mathbf{z}_2, ..., \mathbf{z}_N\}$, or consists of scalars, \textit{e.g.}, $Z=\{z_1, z_2, ..., z_N\}$, and $|Z|$ to denote the size of all the latent variables, \textit{i.e.}, the dimension of the representations.

\vspace{2pt}
\noindent\textbf{Disentanglement Problem:} The independent latent factors and semantic segments can be defined as follows.
\vspace{-5pt}
\begin{definition}[Independent Latent Factors and Segments]
\label{def:1}
Two latent variables $z_i$ and $z_j$ (or semantic segments $\mathbf{z}_i$ and $\mathbf{z}_j$) are independent if the change of the sequential patterns corresponding to variable $z_i$ (segment $\mathbf{z}_i$) is relatively invariant to variable $z_j$ (segment $\mathbf{z}_j$) and vice versa, denoted as $z_{i}  \Perp z_{j}$ ($\mathbf{z}_{i}  \Perp \mathbf{z}_{j}$).
\vspace{-5pt}
\end{definition}

Based on Definition~\ref{def:1}, we formally define the multi-level time-series disentanglement problem 
as follows.


\begin{itemize}[leftmargin=*]
    \item \emph{Individual Latent Factor Disentanglement}:
We aim to learn a set of decomposed  latent variables $Z=\{z_1, z_2, ..., z_N\}$, such that all the pairs of latent variables in $Z$ are independent, \textit{i.e.}, $z_i  \Perp z_j,\forall i,j$, where $i \in \{1, 2,..., N\}$, $j \in \{1, 2, ..., N\}$, and $i \ne j$.

\item \emph{Group Segment Disentanglement}:
We aim to learn a set of decomposed semantic segments $Z=\{\mathbf{z}_1, \mathbf{z}_2, ..., \mathbf{z}_N\}$, such that all the pairs of segments in $Z$ are independent, \textit{i.e.}, $\mathbf{z}_i  \Perp \mathbf{z}_j,\forall i,j$, where $i \in \{1, 2,..., N\}$, $j \in \{1, 2,..., N\}$, and $i \ne j$.
\end{itemize}
\vspace{-5pt}
\subsection{$\beta$-VAE and KL Vanishing Problem} 
In this section, we focus on the latent codes disentanglement task using the generative model family. 
Generative models have shown superior performance in learning the complex distributions. LSTM-VAE~\cite{chung2015recurrent} is widely used in time-series analytics. A latent variable generative model defines a joint distribution between a feature space $Z \in \mathcal{Z},$ and the observation space $\mathbf{x}_{1:T} \in \mathcal{X}$. Suppose we have a simple prior distribution $p(Z)$ is placed over the latent variables. The distribution of the observations conditioned on latent variables is modeled with a deep network $p_{\theta}(\mathbf{x}_{1:T} \mid Z)$. If we have the true underlying distribution as $p_{\mathcal{D}}(\mathbf{x})$ (empirically approximated through the training set), then a natural training objective is to maximize the marginal likelihood as:
\begin{equation}
\mathbb{E}_{p_{\mathcal{D}}(\mathbf{x})}\left[\log p_{\theta}(\mathbf{x})\right]=\mathbb{E}_{p_{\mathcal{D}}(\mathbf{x}_{1:T})}\left[\log \mathbb{E}_{p(Z)}\left[p_{\theta}(\mathbf{x}_{1:T} \mid Z)\right]\right].
\end{equation}
Directly optimizing the likelihood is generally intractable, since computing $p_{\theta}(\mathbf{x})=\int_{z} p_{\theta}(\mathbf{x} \mid Z) p(Z) \mathrm{d} z$ requires integration. A widely-used solution~\cite{kingma2013auto} is to approximate the posterior via an amortized inference distribution $q_{\phi}(Z \mid \mathbf{x})$ and jointly optimize a lower bound to the log likelihood as:
\begin{equation}
\small
\begin{aligned}
\mathcal{L}_{\mathrm{ELBO}}(\mathbf{x}) &=-D_{\mathrm{KL}}\left(q_{\phi}(Z | \mathbf{x}_{1:T}) \| p(Z)\right)+\mathbb{E}_{q_{\phi}(Z | \mathbf{x}_{1:T})}\left[\log p_{\theta}(\mathbf{x}_{1:T} | Z)\right] \\
& \leq \log p_{\theta}(\mathbf{x}_{1:T}).
\label{eq:2}
\end{aligned}
\end{equation}
A disentangled representation~\cite{burgess2018understanding} can be defined as one where single latent units are sensitive to changes in single generative factor, while being relatively invariant to changes in other factors. $\beta$-VAE~\cite{burgess2018understanding} is a variant of the variational autoencoder that attempts to learn a disentangled representation by optimizing a heavily penalized objective with $\beta>1$. The penalization is capable of obtaining models exhibiting disentangled effects in image datasets. With increasing $\beta$, the latent codes could be more disentangled by allowing distributions in the latent space to be further away from each other by less satisfying the KL-term constraint to fit the marginally Gaussian distribution. The semantically similar observations might be mapped closer, thus creating a sort of clusters for interpretations.
\begin{equation}
\small
\begin{aligned}
\mathcal{L}_{\mathrm{\beta-ELBO}}(\mathbf{x})
&=-\beta D_{\mathrm{KL}}\left(q_{\phi}(Z | \mathbf{x}_{1:T}) \| p(Z)\right)+\mathbb{E}_{q_{\phi}(Z | \mathbf{x}_{1:T})}\left[\log p_{\theta}(\mathbf{x}_{1:T}|Z)\right] \\
& \leq \log p_{\theta}(\mathbf{x}_{1:T}).
\label{eq:3}
\end{aligned}
\end{equation}

However, for sequential time-series data, prior work~\cite{wang2019riemannian} shows that: the LSTM generative model often has a strong expressiveness. The reconstruction term in the objective will dominate the KL divergence term during the training phase. 
In this case, the model will generate time-series without effectively making use of latent codes. Besides, the latent variable $Z$ becomes independent from observations $\mathbf{x}$, when the KL divergence term collapses to zero. As a result, the latent variable $Z$ can not serve as an effective representation for the input $\mathbf{x}$, and it can not be used for downstream tasks such as classification and clustering. It can be further referred to as the \textit{information preference} problem~\cite{chen2016variational}. Thus, pushing Gaussian clouds away from each other in the latent space becomes meaningless if latent distributions are unhooked with the observation space.

\vspace{-5pt}
\subsection{Individual Latent Factor Disentanglement}

To alleviate the KL vanishing problem, as well as to preserve the disentangle property, in this section, we introduce the mutual information maximization term to the ELBO decomposition. The goal is to learn a better representation $Z$ that could more effectively capture the semantic characteristics of the input $\mathbf{x}$.
\vspace{-5pt}
\subsubsection{ELBO TC-Decomposition}
To further understand the internal mechanism within the disentangle strategy, we decompose the evidence lower bound (ELBO) to find evidence linking factorial representations to disentanglement. 

Mathematically, the KL term in Eq.~\ref{eq:2} and~\ref{eq:3} can be decomposed with a factorized $p(Z)$ as:
\begin{equation}
\small
\begin{aligned}
& D_{\mathrm{KL}}(q({Z} \mid \mathbf{x}_{1:T})|| p({Z}))=\underbrace{\mathrm{KL}(q(Z, \mathbf{x}_{1:T})|| q(Z) p(\mathbf{x}_{1:T}))}_{\text {(i) Index-Code MI }}\\& \qquad \qquad +\underbrace{\mathrm{KL}(q(Z) \| \prod_{j} q\left(z_{j}\right))}_{\text {(ii) } \text { Total Correlation }} +\underbrace{\sum_{j} \mathrm{KL}\left(q\left(z_{j}\right) \| p\left(z_{j}\right)\right)}_{\text {(iii) Dimension-wise KL }}
\end{aligned}
\label{eq:decompose}
\vspace{-5pt}
\end{equation}
where $z_{j}$ denotes the $j$th dimension of the latent variable. The proof can be found in Appendix A.1. 

The first term can be interpreted as the index-code mutual information (MI). The index-code MI is the mutual information $I_{q_{\phi}}(Z ; \mathbf{x})$ between the data variable and latent variable based on the empirical data distribution $q_{\phi}(Z, \mathbf{x})$. 
From the information theory, the second term is referred to as the total correlation (TC). TC acts as one of many generalizations of mutual information to more than two random variables~\cite{watanabe1960information}. TC is also to measure the dependencies between the variables. The penalty on TC forces the model to find statistically independent factors in the data distribution. A heavier penalty on this term induces a more disentangled representation, and that the existence of this term is the reason $\beta$ -VAE has been successful. One recent study empirically verifies~\cite{chen2018isolating, zhao2019infovae} that $\mathrm{TC}$ is the most important term in this decomposition for learning disentangled representations by only penalizing on this term. The last term is referred to as the dimension-wise KL. This term mainly prevents individual latent dimensions from deviating too far from their corresponding priors. It plays a role as a complexity penalty on the aggregate posterior, which follows from the minimum description length~\cite{hinton1994autoencoders} formulation of the ELBO.

By decomposing the ELBO into separate components, we can have a new perspective for the problem when the latent codes are becoming independent from observations. Introducing a heavier penalty on the ELBO tends to neglect the mutual information between $Z$ and $\mathbf{x}$. Thus, the mutual information becomes vanishingly small. Intuitively, the reason is that the learned $p_{\theta}(\mathbf{x}|Z)$ is the same for all $z \in \mathcal{Z}$, implying that the $Z$ is not dependent on the observations $\mathbf{x}$. This is undesirable because a major goal of unsupervised learning is to learn meaningful latent features that should depend on the observations. If we directly apply $\beta$-VAE models to time-series data for disentanglement, $\beta$-VAE can not effectively trade-off weighting of the mutual information and the reconstruction. Desired disentangled representations emerge when the right balance is found between information preservation (reconstruction cost as regularisation) and latent channel capacity restriction (when $\beta > 1$). Increasing the $\beta$ may intensify the mutual information vanishing problem: when the model has a better quality of disentanglement within the learned latent representations, it penalizes the mutual information simultaneously. This, in turn, can lead to under-fitting or ignoring the latent variables.

 


\vspace{-3pt}
\subsubsection{ELBO DTS-Decomposition} 
We now introduce how to balance the preference between correct inference and fitting data distribution, and provide a new ELBO decomposition solution that sheds light on solving the information preference problem.

Previously, we discussed that optimizing the ELBO tends to push the probability mass of $q_{\phi}(Z \mid \mathbf{x})$ too far from 0. This tendency is a property of the ELBO objective and true for any ${x}$ and ${Z}$. However, this is made worse by the fact that ${x}$ is often higher dimensional compared to ${Z}$, so any error in fitting ${x}$ will be magnified compared to ${Z}$.
To encourage the model to use the latent codes, we add a mutual information maximization term, which encourages a high mutual information between $\mathbf{x}$ and $Z$. In other words, we can address the information preference problem through balancing the preference between correct inference and fitting data. Comparing with the ELBO in LSTM-VAE (in Eq.~\ref{eq:2}), we can rewrite it as: 
\begin{equation}
\vspace{-1pt}
\begin{aligned}
\mathcal{L}_{\mathrm{ELBO}}(\mathbf{x}) &=- D_{\mathrm{KL}}\left(q_{\phi}(Z \mid \mathbf{x}_{1:T}) \| p(Z)\right) + \alpha I_{q_{\phi}}(\mathbf{x}_{1:T} ; Z) \\
& +\mathbb{E}_{q_{\phi}(Z \mid \mathbf{x}_{1:T})}\left[\log p_{\theta}(\mathbf{x}_{1:T} \mid Z)\right]
\end{aligned}
\end{equation}
where $I_{q_{\phi}}(\mathbf{x} ; Z)$ denotes the mutual information between $\mathbf{x}$ and $Z$ under the distribution $q_{\phi}(\mathbf{x} ; Z)$.
But the objective can not be directly optimized. Thus, we rewrite it into another equivalent form:
\begin{equation}
\small
\begin{aligned}
\mathcal{L}_{\mathrm{ELBO}}(\mathbf{x}) &=- D_{\mathrm{KL}}\left(q_{\phi}(Z \mid \mathbf{x}_{1:T}) \| p(Z)\right) +\alpha D_{\mathrm{KL}}\left(q_{\phi}(Z) \| p(Z)\right)\\
& +\mathbb{E}_{q_{\phi}(Z \mid \mathbf{x}_{1:T})}\left[\log p_{\theta}(\mathbf{x}_{1:T} \mid Z)\right].
\end{aligned}
\label{eq:elbotcdecom}
\end{equation}

We find that the mutual information maximization term (the second part of Eq. ~\ref{eq:elbotcdecom}) plays the same role as the first term in the ELBO-TC decomposition (as shown in Eq.~\ref{eq:decompose}). But the optimization directions are contrary. Thus, increasing the disentanglement degree may intensify the KL vanishing problem, and vice versa. To enforce the model to preserve the disentangle property while alleviating the KL vanishing, here, we combine the mutual information regularizer term with the ELBO-TC decomposition in Eq.~\ref{eq:decompose} and merge the mutual information maximization term, then the ELBO can be written as:
\begin{equation}
\small
\begin{aligned}
\mathcal{L}_{\mathrm{ELBO}}&(\mathbf{x}) =- \beta D_{\mathrm{KL}}(q(Z) \| \prod_{j} q\left(z_{j}\right))- \beta \sum_{j} D_{\mathrm{KL}}\left(q\left({z}_{j}\right) \| p\left({z}_{j}\right)\right) \\&+(\alpha-\beta) D_{\mathrm{KL}}\left(q_{\phi}(Z) \| p(Z)\right) +\mathbb{E}_{q_{\phi}(Z \mid \mathbf{x}_{1:T})}\left[\log p_{\theta}(\mathbf{x}_{1:T} \mid Z)\right],
\end{aligned}
\end{equation}
where the mutual information regularizer merges with the first term in the decomposition analysis. Mathematically, we alleviate the KL vanishing problem by introducing the mutual information maximization term, while preserve a heavier penalty (when $\beta > 1$) on the total correlation and the dimension-wise KL to keep the disentangle property. 

\vspace{-5pt}
\subsection{Latent Group Segment Disentanglement}

The individual factor disentanglement introduced in the last section can't guarantee interpretability. For time-series applications, it's hard to interpret the single latent factor, and human-interpretable semantic concepts typically rely on multiple latent factors instead of a single factor. Thus, in this section, we introduce how to disentangle latent codes of time-series into group segments (Section~\ref{subsub:group}). We further show that our proposed group segment disentanglement could benefit the domain adaptation task (Section~\ref{subsub:domain}).


\subsubsection{Group Segment Disentanglement}
\label{subsub:group}
Different from individual variable analysis, in this section, we aim to learn decomposed semantic segments that contain batches of latent variables. One main advantage of the group segment disentanglement is to extract more than one latent factors, which simultaneously contribute to one complex sequential semantic concept. 

Assume that the semantic segments are independent factors, \textit{i.e.}, $\mathbf{z}_{m} \Perp \mathbf{z}_{n}$ with every pairs of $m, n$ sampled from $\mathbf{z} = \{\mathbf{z}_{1}...\mathbf{z}_{k} \}$, where $k$ is the number of the segments. For ease of presentation, we take $m, n$ as two semantic segments for illustration. Our method could be easily extended to multiple segments scenarios. Given $\mathbf{x}$, the new time-series is generated from two independent latent group segments, \textit{i.e.}, $\mathbf{z}_{m}$ encodes the $m_{th}$-segment and $\mathbf{z}_{n}$ encodes the $n_{th}$-segment.
\begin{theorem}
\vspace{-2pt}
Assume that two group segments are independent, \textit{i.e.}, $\mathbf{z}_{m} \Perp \mathbf{z}_{n}$, and let $\mathbf{z}=\left\{\mathbf{z}_{m}, \mathbf{z}_{n}\right\}$, the empirical error on the disentangled segments with a hypothesis $h$ is:
\begin{equation}
    \epsilon(h) = \mathrm{E}_{\mathbf{z}_{m} \sim \mathcal{Z}}\left[C_{m}\left(\mathbf{z}_{m}\right)-h\left(\mathbf{z}_{m}\right)\right]  + \mathrm{E}_{\mathbf{z}_{n} \sim \mathcal{Z}}\left[C_{n}\left(\mathbf{z}_{n}\right)-h\left(\mathbf{z}_{n}\right)\right]
\end{equation}
where $\epsilon_{\mathrm{T}}^{\mathrm{y}}(h)$ denotes the empirical error of \model with respect to $h$.
\label{theo:indi}
\end{theorem}
\noindent \textit{Proof}. Since $\mathbf{z}_{m} \Perp \mathbf{z}_{n} $, we can derive the empirical error as follows:
\begin{equation}
\begin{aligned}
\epsilon(h) &=  \mathrm{E}_{(\mathbf{z}_{m}, \mathbf{z}_{n})\sim \mathcal{Z}}\left[C\left(\mathbf{z}\right)-h\left(\mathbf{z}\right)\right]
\\  &= \mathrm{E}_{\mathbf{z}_{m} \sim \mathcal{Z}}\left[C_{m}\left(\mathbf{z}_{m}\right)-h\left(\mathbf{z}_{m}\right)\right]  + \mathrm{E}_{\mathbf{z}_{n} \sim \mathcal{Z}}\left[C_{n}\left(\mathbf{z}_{n}\right)-h\left(\mathbf{z}_{n}\right)\right]. 
\end{aligned}
\end{equation}

Based on the independence property between $\mathbf{z}_{m}$ and $\mathbf{z}_{n},$ the distribution of $\mathcal{Z}$ can be decomposed into two parts so as to the error. Following the evidence lower bound of the marginal likelihood in the Eq.~\ref{eq:elbotcdecom}, we get a similar form for group segments:
\begin{equation}
\small
\begin{aligned}
    & \mathcal{L}_{\mathrm{ELBO-G}}(\mathbf{x}) 
     = -D_{\mathrm{KL}}\left(q_{\phi_m}(\mathbf{z}_m \mid \mathbf{x}_{1:T}) \| p(\mathbf{z}_n)\right)-D_{\mathrm{KL}}\left(q_{\phi_n}(\mathbf{z}_n \mid \mathbf{x}_{1:T}) \| p(\mathbf{z}_n)\right)
    \\
    & \qquad + \mathbb{E}_{q_{\phi_m}(\mathbf{z}_m,\mathbf{z}_n \mid \mathbf{x}_{1:T})}\left[\log p_{\theta}(\mathbf{x}_{1:T} \mid \mathbf{z}_m,\mathbf{z}_n)\right]  +\alpha D_{\mathrm{KL}}\left(q_{\phi}(Z) \| p(Z)\right).
\end{aligned}
\end{equation}


We assume that $P\left(\boldsymbol{z}_{m}\right), P\left(\boldsymbol{z}_{n}\right) \sim \mathcal{N}(\mathbf{0}, \mathbf{I})$, and $\phi_{m}$ and $\phi_{n}$ are the parameters of the encoder. By applying a reparameterization trick, we use sequential models (LSTMs or TCNs~\cite{bai2018empirical}) $H_{m}\left(G(\mathbf{x}); \phi_{y}\right)$ and $H_{n}\left(G(\mathbf{x}); \phi_{d}\right)$ as the universal approximator of $q$ to encode the data into $\mathbf{z}_{m}$ and $\mathbf{z}_{n}$, respectively.

Meanwhile, we want to make the $\mathbf{z}_m$ task-$n$-invariant, and make the $\mathbf{z}_n$ task-$m$ invariant. That is, we want to make the $C_{m}\left(\mathbf{z}_{n};\theta_m\right)$ and $C_{n}\left(\mathbf{z}_{m};\theta_n\right)$ less discriminative. To obtain task-invariant representations, we seek the parameters $\theta_m$ of the feature mapping that maximize the loss of
the $q_{\theta_m}$ (by making the distributions of different  $\mathbf{z}_m$  as close as possible), while simultaneously seeking the parameters  $\theta_n$ that minimize the loss of $\mathbf{z}_m$. We aim to solve:
\begin{equation}
\mathbb{E}_{\text {m}}\left(\phi_{y}, \theta_{m}, \theta_{n}\right) = \mathbb{E}\left(C_{m}\left(\mathbf{z}_{m} ; \theta_{m}\right), y_{m}\right) - \lambda \mathbb{E}\left(C_{n}\left(\mathbf{z}_{m} ; \theta_{n}\right), y_{n}\right).
\label{eq:em}
\end{equation}
Moreover, we seek to minimize the loss of $\mathbf{z}_n$. Similarly, we have:
\begin{equation}
\mathbb{E}_{\text {n}}\left(\phi_{y}, \theta_{m}, \theta_{n}\right) = \mathbb{E}\left(C_{n}\left(\mathbf{z}_{n} ; \theta_{n}\right), y_{n}\right) - \lambda \mathbb{E}\left(C_{m}\left(\mathbf{z}_{n} ; \theta_{m}\right), y_{m}\right).
\label{eq:en}
\end{equation}
From Eq.~\ref{eq:em} and~\ref{eq:en}, we are seeking the parameters $\hat{\theta}_{f}, \hat{\theta}_{y}, \hat{\theta}_{d}$ as:
\begin{equation}
\begin{array}{c}
\left(\hat{\theta}_{f}, \hat{\theta}_{y}\right)=\arg \min _{\theta_{f}, \theta_{y}} E\left(\theta_{f}, \theta_{y}, \hat{\theta}_{d}\right) \\
\hat{\theta}_{d}=\arg \max _{\theta_{d}} E\left(\hat{\theta}_{f}, \hat{\theta}_{y}, \theta_{d}\right),
\end{array}
\end{equation}
where the parameter $\lambda$ controls the trade-off between the two objectives that shape the features during training. The update process is very similar to vanilla stochastic gradient descent (SGD) updates for feed-forward deep models. One thing that needs to point out is the $-\lambda$ factor, which tries to make disentangled features less discriminative for the irrelevant task. Here, we use a gradient reversal layer (GRL)~\cite{ganin2015unsupervised} to exclude the discriminative information. During the forward propagation, GRL acts as an identity transform. During the backpropagation, GRL takes the gradient from the subsequent level, and multiplies the gradient by a negative constant, then passes it to the preceding layer. 

\vspace{-3pt}
\subsubsection{Domain Adaptation as Concrete Example}
\label{subsub:domain}
To further illustrate the benefits of the proposed group segments disentanglement for time-series, we apply it to the domain adaptation problem as a concrete application scenario.


In the unsupervised domain adaptation problem, we use the labeled samples $D_{S}=\left\{\boldsymbol{x}_{1:T}^{S}, y_{i}^{S}\right\}_{i=1}^{n_{S}}$ on the source domain to classify the unlabeled samples $D_{T}=$ $\left\{\boldsymbol{x}_{j}^{T}\right\}_{j=1}^{n_{T}}$ on the target domain. We aim to obtain two independent latent variables with disentanglement, including a domain-dependent latent variable $\mathbf{z}_{d}$ and a class-dependent latent variable $\mathbf{z}_{y}$. These two variables are expected to encode the domain information and the class information, respectively. Then, we can use the class-dependent latent variable for classification since it is domain-invariant. Mathematically, deriving from Theorem ~\ref{theo:indi}, we have: 
\begin{theorem}
\vspace{-2pt}
Assume that the class and the domain factors are independent, \textit{i.e.}, $\mathbf{z}_{y} \Perp \mathbf{z}_{d} .$ Let $\mathbf{z}=\left\{\mathbf{z}_{y}, \mathbf{z}_{d}\right\},$ and
the error on the disentangled source and target domain with a hypothesis $h$ is:
$\begin{aligned}
\centering
\epsilon_{S}(h) &=\mathrm{E}_{\mathbf{z}_{y} \sim \mathcal{Z}_{S}}\left[C\left(\mathbf{z}_{y}\right)-h\left(\mathbf{z}_{y}\right)\right]+\mathrm{E}_{\mathbf{z}_{d} \sim \mathcal{Z}_{S}}\left[C\left(\mathbf{z}_{d}\right)-h\left(\mathbf{z}_{d}\right)\right] \\
\epsilon_{T}(h) &=\mathrm{E}_{\mathbf{z}_{y} \sim \mathcal{Z}_{T}}\left[C\left(\mathbf{z}_{y}\right)-h\left(\mathbf{z}_{y}\right)\right]+\mathrm{E}_{\mathbf{z}_{d} \sim \mathcal{Z}_{T}}\left[C\left(\mathbf{z}_{d}\right)-h\left(\mathbf{z}_{d}\right)\right]. 
\end{aligned}
$
\label{theo:independent}
\end{theorem}

According to the Theorem ~\ref{theo:independent}, we can find that, the disentangled empirical classification error rate with respect to $h$ in the source domain is lower than before disentanglement
($\epsilon_{S}^{y}(h)=\epsilon_{S}(h)-\epsilon_{S}^{d}(h)$, where $\epsilon_{S}^{d}(h) \ge 0$).  Prior work~\cite{cai2019learning} theoretically prove that the disentanglement of the representation space could be helpful and necessary for obtaining a lower classification error rate. Furthermore, a lower classification error rate on the source domain will tighten the error bound at the target domain.

We measure the discrepancy distance between the source and target distribution with respect to hypothesis $h$. Formally, we have:
\begin{equation}
\small
    \begin{aligned}
d_{\mathcal{H} \Delta \mathcal{H}}(\mathcal{S}, \mathcal{T})=2\sup _{h_{1}, h_{2} \in \mathcal{H}} | P_{\mathbf{f} \sim \mathcal{S}}\left[h_{1}(\mathbf{f}) \neq h_{2}(\mathbf{f})\right]
-P_{\mathbf{f} \sim \mathcal{T}}\left[h_{1}(\mathbf{f}) \neq h_{2}(\mathbf{f})\right]|
\end{aligned}
\end{equation}

A probabilistic bound on the performance $\varepsilon_{\mathcal{T}}(h)$ of  classifier $h$ from $\mathcal{T}$ is  evaluated on the target domain, given its performance $\varepsilon_{\mathcal{S}}(h)$ on the source domain, where $\mathcal{S}$ and $\mathcal{T}$ are source and target distributions, respectively:
\begin{equation}
\varepsilon_{\mathcal{T}}(h) \leq \varepsilon_{\mathcal{S}}(h)+\frac{1}{2} d_{\mathcal{H} \Delta \mathcal{H}}(\mathcal{S}, \mathcal{T})+C
\end{equation}
where $C$ does not depend on a particular $h$. Prior work~\cite{ganin2015unsupervised} proves that optimal discriminator gives the upper bound for the $d_{\mathcal{H} \triangle \mathcal{H}}(\mathcal{S}, \mathcal{T})$, so that tighter bound $\varepsilon_{\mathcal{S}}(h)$ from the source domain could lead to a better approximation of $\varepsilon_{\mathcal{T}}(h)$.


\vspace{-2pt}
\section{Why does Proposed Method Work?}
In this section, we introduce intuition as well as theoretical justifications of \modelnospace, and its relationships with existing methods.

\vspace{-5pt}
\subsection{Theoretical Possibility}
One recent work essentially shows that without inductive biases on both models and datasets, the unsupervised disentangle task is fundamentally impossible~\cite{locatello2019challenging}.
\begin{theorem}
\vspace{-2pt}
 Suppose $p(\mathbf{z})$ is a d-dimensional distribution and $d>1$, let $\mathbf{z} \sim P$ denote any distribution which admits a density $p(\mathbf{z})=\prod_{i=1}^{d} p\left(\mathbf{z}_{i}\right).$ Then, there exists an infinite family of bijective functions $f: \operatorname{supp}(\mathbf{z}) \rightarrow$ $\operatorname{supp}(\mathbf{z})$ such that $\frac{\partial f_{i}(\boldsymbol{u})}{\partial u_{i}} \neq 0$ almost everywhere for all
i and $j$ (i.e., $\mathbf{z}$ and $f(\mathbf{z})$ are completely entangled) and $P(\mathbf{z} \leq \boldsymbol{u})=P(f(\mathbf{z}) \leq \boldsymbol{u})$ for all $\boldsymbol{u} \in \operatorname{supp}(\mathbf{z})$ (i.e., they
have the same marginal distribution).~\cite{locatello2019challenging}
\label{theo:1}
\vspace{-0pt}
\end{theorem}
Intuitively, after observing $\mathbf{x}$, we can construct infinitely many generative models, which have the same marginal distribution of $\mathbf{x}$. Any one of these models could be the true causal generative model for the data, and the right model cannot be identified given only the distribution of $\mathbf{x}$.

While Theorem ~\ref{theo:1} shows that unsupervised disentanglement learning is fundamentally impossible for arbitrary generative models, it is still theoretically possible for our proposed \modelnospace. First, from the model structure perspective, our proposed \model exploits TCN and LSTM-like models, which impose the inductive biases towards preserving contextual information over long sequences in the design of the neural network architecture. Second, from the data structure perspective, time-series as sequential data, contain trend and seasonal patterns, with the serial correlation between subsequent observations. Third, from the implicit supervision perspective, \model could incorporate additional weak supervision (like signals from the source domain) to guide a better disentangled representation. Thus, we make explicit and implicit inductive bias available during the training phase.
\vspace{-3pt}
\subsection{Relationship to Existing Models}
Here, we discuss the relationship of DTS to existing models based on the information bottleneck principle~\cite{tishby2015deep}. The proposed \modelnospace, including individual latent factors and group segments, can be considered as a product of variational decomposition of mutual information terms in the information bottleneck (IB) framework. 
\vspace{-2pt}
\subsubsection{Information Bottleneck for Unsupervised Models} 
The unsupervised IB can be treated as a compression process from $\mathbf{x}$ to $\mathbf{z}$ via the parametrized mapping $q_{\phi}(\mathbf{z} \mid \mathbf{x})$. This process leads to a bottleneck representation $\mathbf{z}$ yet preserving a certain level of information $I_{x}$ in $\mathbf{z}$ about $\mathbf{x}$. Accordingly, this problem can be formulated as:
\begin{equation}
    \min _{\phi: I(\mathbf{Z} ; \mathbf{X}) \geq I_{x}} I_{\boldsymbol{\phi}}(\mathbf{X} ; \mathbf{Z}),
\end{equation}
and in the Lagrangian formulation as a minimization of:
\begin{equation}
    \mathcal{L}(\boldsymbol{\phi})=I_{\boldsymbol{\phi}}(\mathbf{X} ; \mathbf{Z})-\beta I(\mathbf{Z} ; \mathbf{X}).
\end{equation}

\vspace{-5pt}
\subsubsection{Relation to LSTM-VAE and $\beta$-VAE}  Encoders in LSTM-VAE maps a data point from the observation space into a probabilistic output of Gaussian cloud with mean $\mu(\mathbf{x})$ and `ellipsoid' orientation determined by the diagonal covariance matrix $\operatorname{diag}(\sigma(\mathbf{x}))$.  Comparing with LSTM-VAE, $\beta$-VAE relaxes the stochastic `compression' via mapping everything to a Gaussian heap by applying the relaxation parameter that might give more preference to the reconstruction loss. Our proposed \model can be considered as  another form of compression by the minimization of $I_{\boldsymbol{\phi}}(\mathbf{X}; \mathbf{Z})$.  \model relaxes the condition to map all conditional distributions to one Gaussian heap.

\vspace{-5pt}
\subsubsection{Relation to Domain Adaptation Methods} Existing domain adaptation methods focus on (i) utilizing maximum mean discrepancy to measure the domain alignment~\cite{tzeng2014deep}; and (ii) extracting the domain-invariant representation as transferable common knowledge on the feature space~\cite{cai2019learning,wilson2020multi}. Motivated by the success of disentanglement in the image domain, \model aims to extract the group segments in the latent disentangled semantic representation of the data. \model also introduces interpretability in the latent space via weak-supervised signals as imposing special constraints on the latent codes.
\vspace{-5pt}
\subsubsection{Complexity Analysis} 
\model can disentangle latent factors for sequential data, scale to complicated datasets, and typically requires no more training time than vanilla VAEs. It does not require an exponentially growing computational cost in the number of factors. 


\vspace{-5pt}
\section{Experiments}
We conduct experiments to answer research questions as follows:

\vspace{-2pt}
\begin{itemize}[leftmargin=*]

\item \textbf{Q1}: Compared with non-disentangled methods, how quantitatively effective is the proposed disentangled strategy? 

\item \textbf{Q2}: Does individual latent factor disentanglement benefit the generation process of the time-series in a more informative way?

\item \textbf{Q3}: Whether or not latent group segment disentanglement strategy could separate semantic concepts? 
\end{itemize}

\vspace{-5pt}
\subsection{Experiment Setup}

In this section, we introduce benchmark datasets and baselines. For hyperparameter settings and network configurations, please refer to Appendix A.3 for the details.
\vspace{-5pt}
\subsubsection{Datasets}
We evaluate \model on five benchmark datasets for individual latent factor disentanglement and group segment disentanglement with domain adaptation as a concrete task. For the time-series representation task, it is label-free. For the domain adaptation task, the training-test split is 80\% and 20\% respectively, and the training data is further split into training-validation with the same proportions.  We segment the series into non-overlapping windows of 128 time steps (more details in Appendix A.2). 
\begin{itemize}[leftmargin=*]
\item \textbf{Human Activity Recognition (HAR)}~\cite{anguita2013public}: contains sequential accelerometer, gyroscope, and estimated body acceleration data from $30$ participants at 50 Hz.
\item \textbf{Heterogeneity Human Activity Recognition (HHAR)}~\cite{stisen2015smart}: includes accelerometer data from $31$ smartphones of different manufacturers and models positioned in various orientations. 
\item \textbf{WISDM Activity Recognition (WISDM AR)}~\cite{kwapisz2011activity}: contains $33$ participants’ accelerometer data, which are sampled at 20 Hz.
\item \textbf{uWave}~\cite{liu2009uwave}: is a
large gesture library with over $4000$ samples collected from eight users over an elongated period of time for a gesture vocabulary with eight gesture patterns.
\item \textbf{ECG Signal}~\cite{blackburn1960electrocardiogram}: contains heartbeats annotated by at least two cardiologists. The annotations are mapped into $5$ groups in accordance with the AAMI standard. 
\end{itemize}

\vspace{-5pt}
\subsubsection{Baselines}
We compare \model with three state-of-the-art domain adaptation algorithms and one time-series generative model:
\begin{itemize}[leftmargin=-4pt]
\item \textbf{Recurrent Domain Adversarial Neural Network(R-DANN)}~\cite{ajakan2014domain}: employs an LSTM network, and promotes the emergence of features that are (i) discriminative for the learning task on the source domain and (ii) indiscriminate with respect to the shift between the domains.
\item \textbf{Variational Recurrent Adversarial Deep Domain Adaptation (VRADA)}~\cite{purushotham2017variational}: uses a variational RNN and trains adversarially to capture  temporal relationships that are domain-invariant.
\item \textbf{Convolutional Deep Domain Adaptation (CoDATS)}~\cite{wilson2020multi}: leverages domain-invariant domain adaptation methods to operate on time-series data, and utilizes weak supervisions from labels.
\item \textbf{LSTM-VAE}~\cite{chung2015recurrent}: 
is similar to an auto-encoder. It learns an LSTM as the encoder that maps the sequential data to a latent representation in a probabilistic manner, and decodes the latents back to data.

\end{itemize}

\subsection{Performance Evaluation}
To answer the research question \textbf{Q1},  we apply \model to domain adaptation tasks to validate the \emph{effectiveness of the group segmentation} (Section~2.4), and compare \model with the state-of-the-art algorithms.
During the training phase, only the training dataset is accessible: labeled data as the source domain and unlabeled data as the target domain. During the evaluation, the test data becomes available as the target domain for quantitative evaluations and visualizations.
\begin{table}[!t]
\vspace{-10pt}
\small
\setlength{\tabcolsep}{1.2mm}{
\begin{tabular}{@{}c|cccccc@{}}
\toprule
Problem          & W/O  & R-DANN & VRADA & CoDATS & \model    & Target \\ \midrule
HAR 2 → 11       & 83.3           & 80.7   & 64.1  & 74.5   & \textbf{84.3}  & 100.0           \\
HAR 7 → 13       & 89.9           & 75.3   & 78.3  & 96.5   & \textbf{98.1}  & 100.0           \\
HAR 9 → 18       & 31.1           & 56.6   & 59.8  & 85.8   & \textbf{89.8}  & 100.0           \\
HAR 14 →19       & 62.0           & 71.3   & 64.4  & 98.6   & \textbf{100.0} & 100.0           \\
HAR 18 →23       & 89.3           & 78.2   & 72.9  & 89.3   & \textbf{94.9}  & 100.0           \\
HAR 7 → 24       & 94.4           & 84.8   & 93.9  & 99.1  & \textbf{100.0} & 100.0           \\
HAR 17 →25       & 57.3           & 66.3   & 52.0  & 97.6   & \textbf{100.0} & 100.0           \\
\hline
HHAR 1 →3        & 77.8           & 85.1   & 81.3  & 90.8   & \textbf{93.7}  & 99.2            \\
HHAR 3 →5        & 68.8           & 85.4   & 82.3  & 94.3   & \textbf{95.9}  & 99.0            \\
HHAR 4 →5        & 60.4           & 70.4   & 71.6  & 94.2   & \textbf{94.9}  & 99.0            \\
HHAR 1 →6        & 72.1           & 81.7   & 74.9  & 90.8   & \textbf{92.1}  & 98.8            \\
HHAR 4 →6        & 48.0           & 64.6   & 62.7  & 85.3   & \textbf{92.3}  & 98.8            \\
HHAR 5 →6        & 65.1           & 54.4   & 60.0  & 91.7   & \textbf{92.5}  & 98.8            \\
HHAR 5 →8        & 95.3           & 82.5   & 87.5  & 95.8   & \textbf{97.9}  & 99.3            \\
\hline
WISDM 4 → 15   & 78.2           & 69.2   & 82.7  & 81.4   & \textbf{82.9}  & 100.0           \\
WISDM 2 → 25   & 81.1           & 57.8   & 72.2  & 90.6   & \textbf{95.8}  & 100.0           \\
WISDM 25 → 29  & 47.1           & 61.6   & 81.9  & 74.6   & \textbf{82.2}  & 95.7            \\
WISDM 7 → 30   & 62.5           & 41.7   & 61.9  & 73.2   & \textbf{89.2}  & 100.0           \\
WISDM 21 → 31  & 57.1           & 61.0   & 68.6  & 92.4   & \textbf{96.4}  & 97.1            \\
WISDM 2 → 32   & 60.1           & 49.0   & 66.7  & 68.6   & \textbf{70.7}  & 100.0           \\
WISDM 1 → 7    & 68.5           & 44.8   & 63.0  & 66.1   & \textbf{72.7}  & 96.4            \\
\hline
uWave 2 → 5      & 86.3           & 33.3   & 18.5  & 98.2   & \textbf{100.0} & 100.0           \\
uWave 3 → 5      & 82.7           & 63.7   & 32.4  & 92.9   & \textbf{95.6}  & 100.0           \\
uWave 2 → 6      & 86.0           & 34.5   & 25.3  & 93.8   & \textbf{97.8}  & 100.0           \\
uWave 2 → 7      & 85.1           & 53.9   & 12.2  & 91.4   & \textbf{98.9}  & 100.0           \\
uWave 3 → 7      & 95.5           & 64.0   & 30.4  & 92.0   & \textbf{98.9}  & 100.0           \\
uWave 1 → 8      & 100.0          & 78.6   & 11.0  & 93.8  & \textbf{100.0} & 100.0           \\
uWave 7 → 8      & 95.2           & 49.7   & 12.5  & 93.8   & \textbf{96.7}  & 100.0           \\
\bottomrule
\end{tabular}}
\caption{Target classification accuracy (based on the class-dependent representation $\mathbf{z}_y$) for time-series domain adaptation (from source to target) on randomly-chosen problems for each dataset, adapting between different users. We include no adaptation as an approximate lower bound (W/O), and models trained directly on labeled target data as
upper bound (Target) (more results are in Appendix A.4).}
\label{tab:domainadaptation}
\vspace{-10pt}
\end{table}

\begin{table}[t]
\vspace{-5pt}
\small
\setlength{\tabcolsep}{1mm}{
\begin{tabular}{@{}ccc|cccc@{}}
\toprule
                  Dataset  & \textbf{Acc/D-S} & \textbf{Acc/D-T} & \textbf{Acc/C-S} & \textbf{Acc/C-T} & \textbf{AUC/C-S} & \textbf{AUC/C-T} \\ \midrule
HAR     & 100.0                      & 100.0                      & 100.0                    & 97.2                     & 100                 & 99.2                \\
HAR/r   & 54.7                       & 66.7                       & 35.9                     & 22.2                     & 54.6                & 53.7                \\
\hline
HHAR    & 96.9                       & 98.8                       & 97.9                     & 91.1                     & 99.8                & 98.7                \\
HHAR/r  & 70.8                       & 70.9                       & 30.5                     & 17.6                     & 58.3                & 52.7                \\
\hline
WISDM   & 100                        & 100                        & 94.5                     & 70.7                     & 99.0                & 85.9                \\
WISDM/r & 67.3                       & 46.7                       & 38.2                     & 33.3                     & 69.7                & 67.9                \\
\hline
uWave      & 100                        & 100                        & 99.1                     & 94.4                     & 99.0                & 85.9                \\
uWave/r    & 55.4                       & 48.2                       & 18.8                     & 12.5                     & 54.6                & 53.7                \\ \bottomrule
\end{tabular}} 
\caption{Ablation studies: discriminability. The results are listed as X-Y (X: D denotes discriminability  on domains, C denotes discriminability on tasks; Y: S and T denote features sampled from source and target domains, respectively). Top rows: performance based on the disentangled domain-dependent $\mathbf{z}_d$, or class-dependent $\mathbf{z}_y$ for time-series domain adaptation. Bottom rows(/r): performance based on the domain-invariant $\mathbf{z}_y$ and class-invariant $\mathbf{z}_d$ segments.}
\label{tab:ablation}
\vspace{-15pt}
\end{table}
\vspace{-5pt}
\subsubsection{Quantitative Results} 


Table~\ref{tab:domainadaptation} compares the performance of \model and the baselines on HAR, HHAR, WISDM AR, and uWave datasets. We include no adaptation as an approximate lower
bound, and
models trained directly on labeled target data as
upper bound. After the group disentanglement, two variables are used to encode the domain information and the class information, respectively. We use the class-dependent latent variable for classification since it is domain-invariant. We observe that \model outperforms the baselines with consistently $+3\%$ higher accuracy over all datasets. 
These results ascertain the effectiveness of ~\model in boosting the performance of domain adaptation by obtaining domain-invariant transferable components common knowledge. The enhanced results also validate eliminating irrelevant information from group  disentanglement could prevent negative transferring.

\begin{figure}[t]
\vspace{-12pt}
        \begin{subfigure}[b]{0.45\textwidth}
         \includegraphics[width=\linewidth]{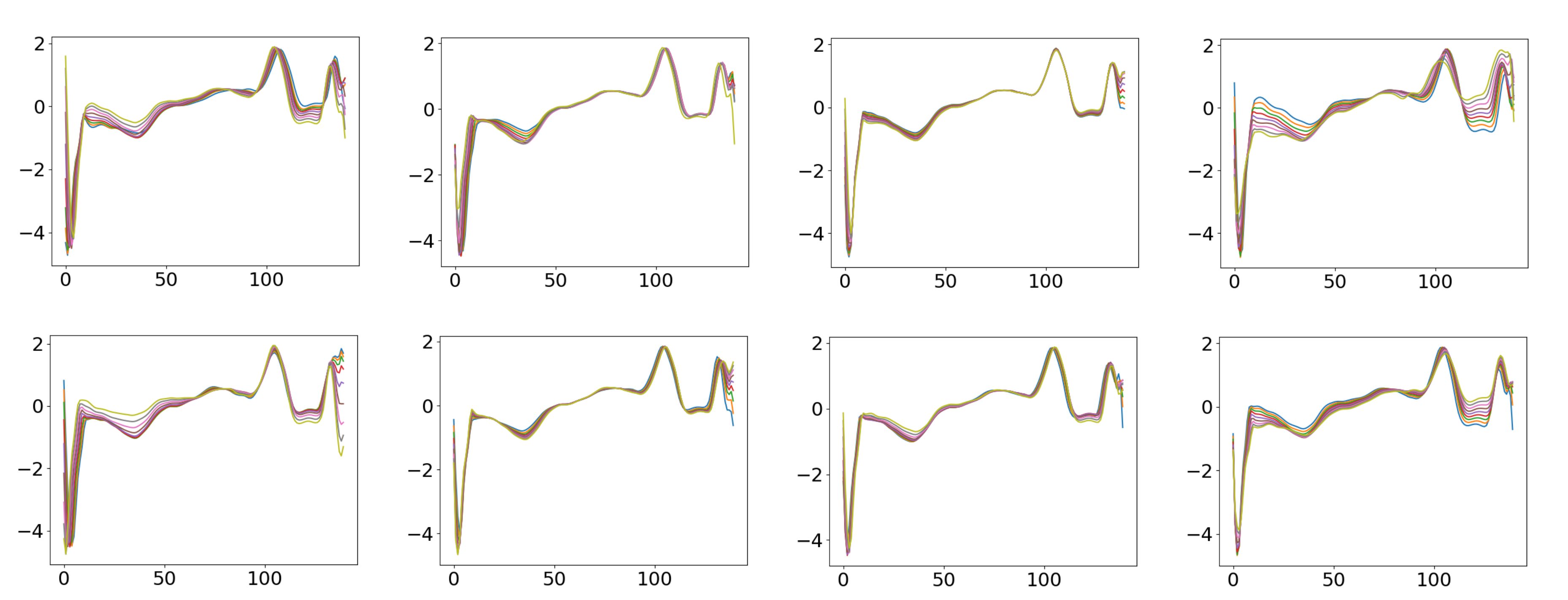}
        \end{subfigure}%
\hrule
        \begin{subfigure}[b]{0.45\textwidth}
 \includegraphics[width=\linewidth]{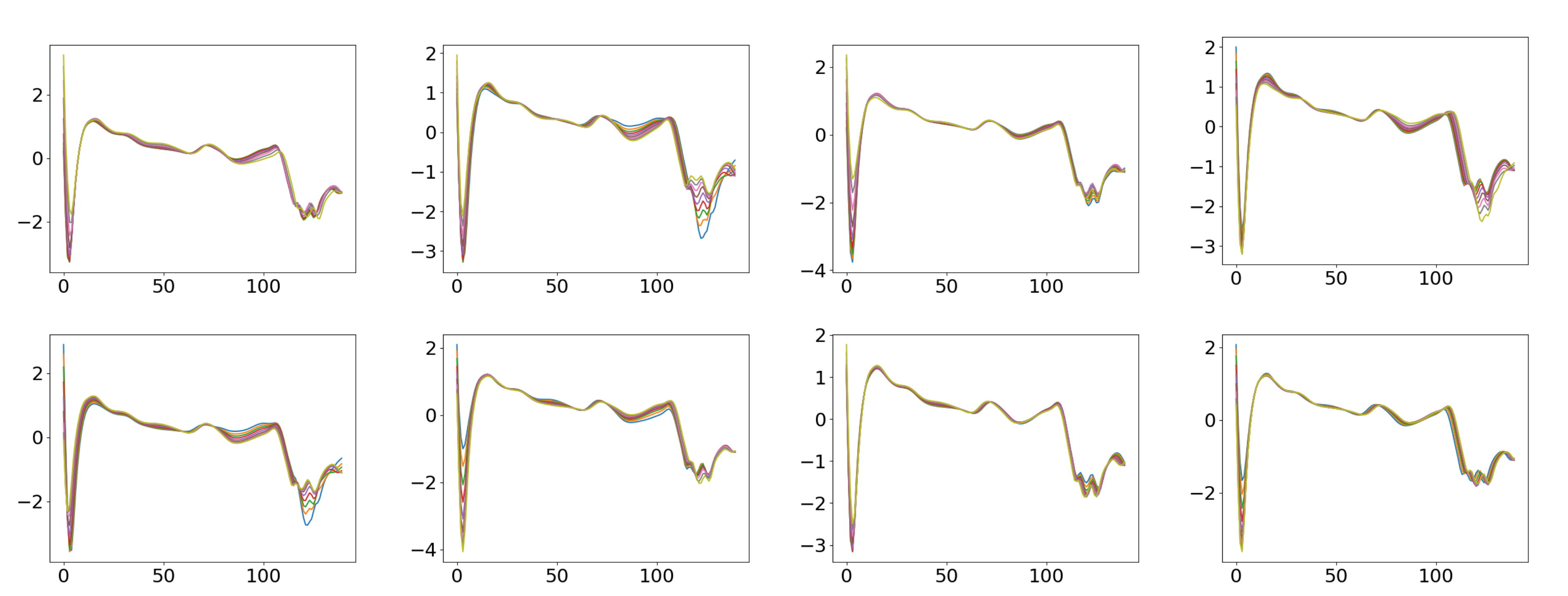}
                \caption{Visualizing the generated time-series from LSTM-VAE. }
        \end{subfigure}%
        \newline
        \begin{subfigure}[b]{0.45\textwidth} \includegraphics[width=\linewidth]{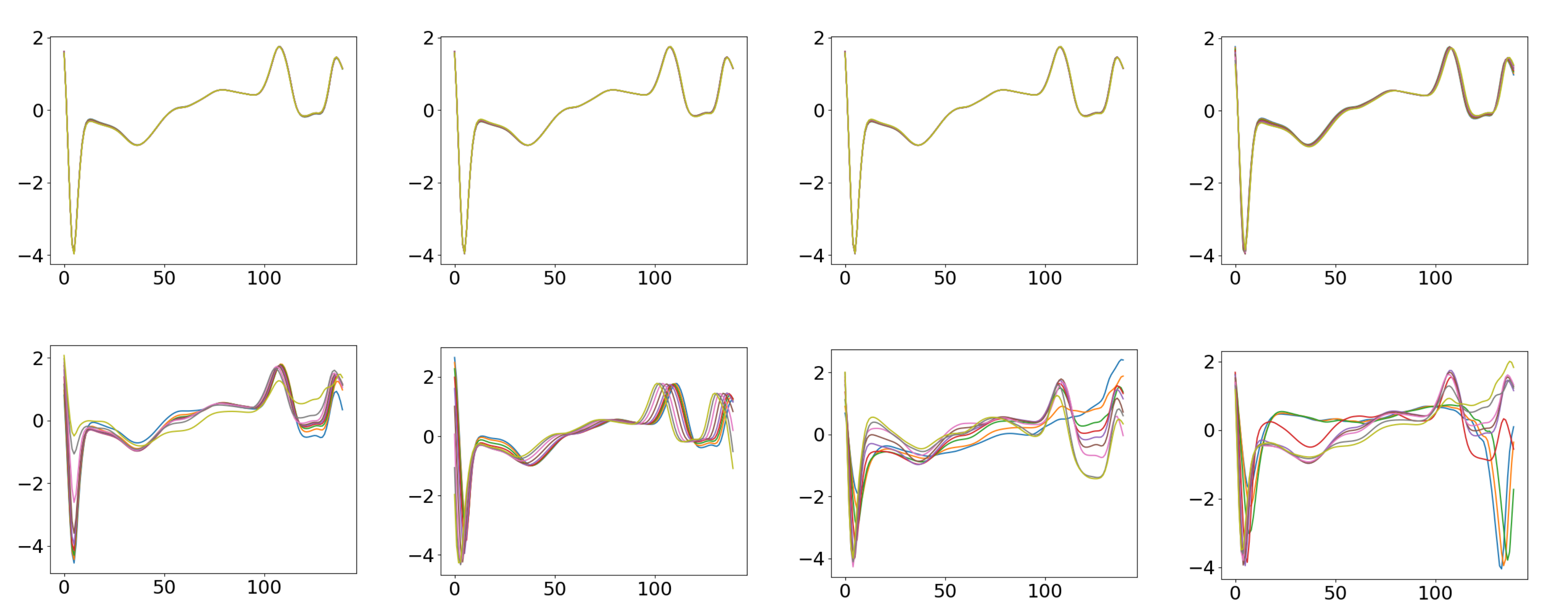}
        \end{subfigure}%
\hrule
        \begin{subfigure}[b]{0.45\textwidth}
                \includegraphics[width=\linewidth]{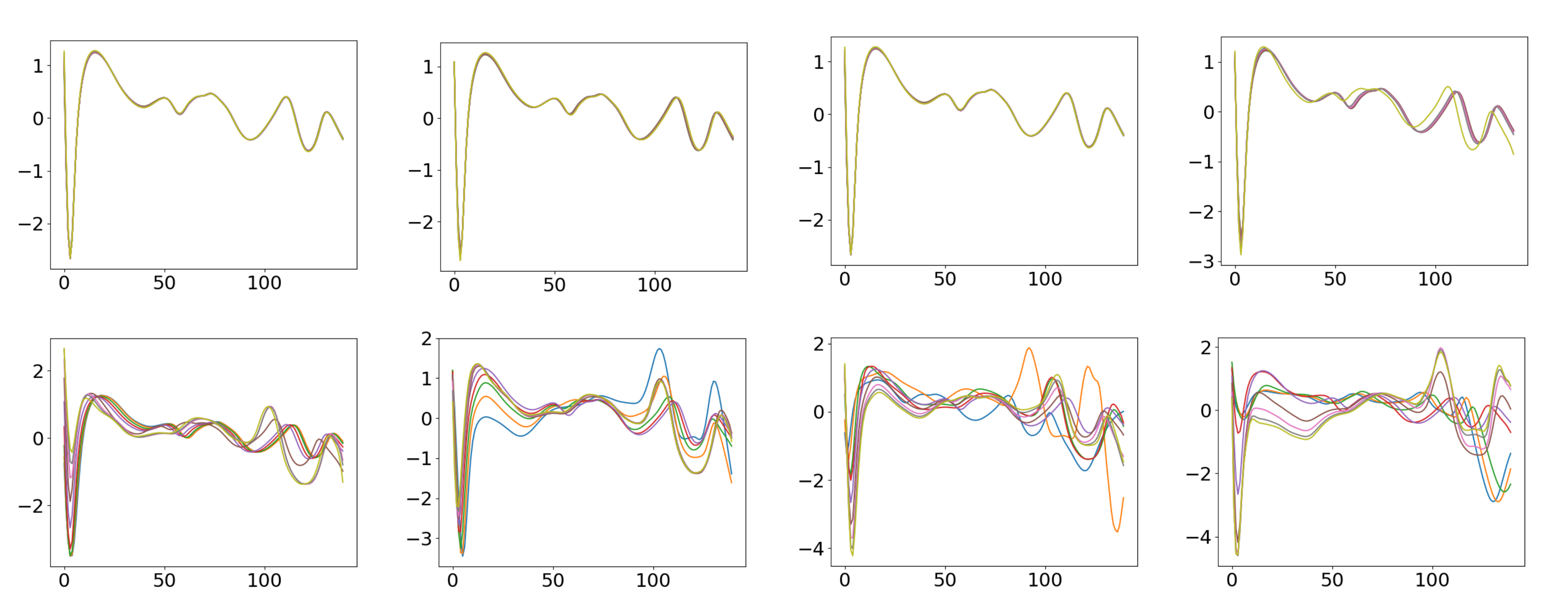}
                \caption{Visualizing the generated time-series from \modelnospace.}
        \end{subfigure}%
        \vspace{-9pt}
        \caption{
        Comparison of learned latent variables. Traversals depict an interpretable property in generating time-series from eight-dimensions of the latent codes $Z$ (shown as 8 subfigures in a group). Traversals are sampled from two different time-series of ECG (separated by the black line), with range [-4,4] (shown as 9 lines in one subfigure). }\label{fig:latent}
        \vspace{-20pt}
\end{figure}

\begin{figure*}[t]
        \vspace{-25pt}
    \centering
    \includegraphics[width=\linewidth]{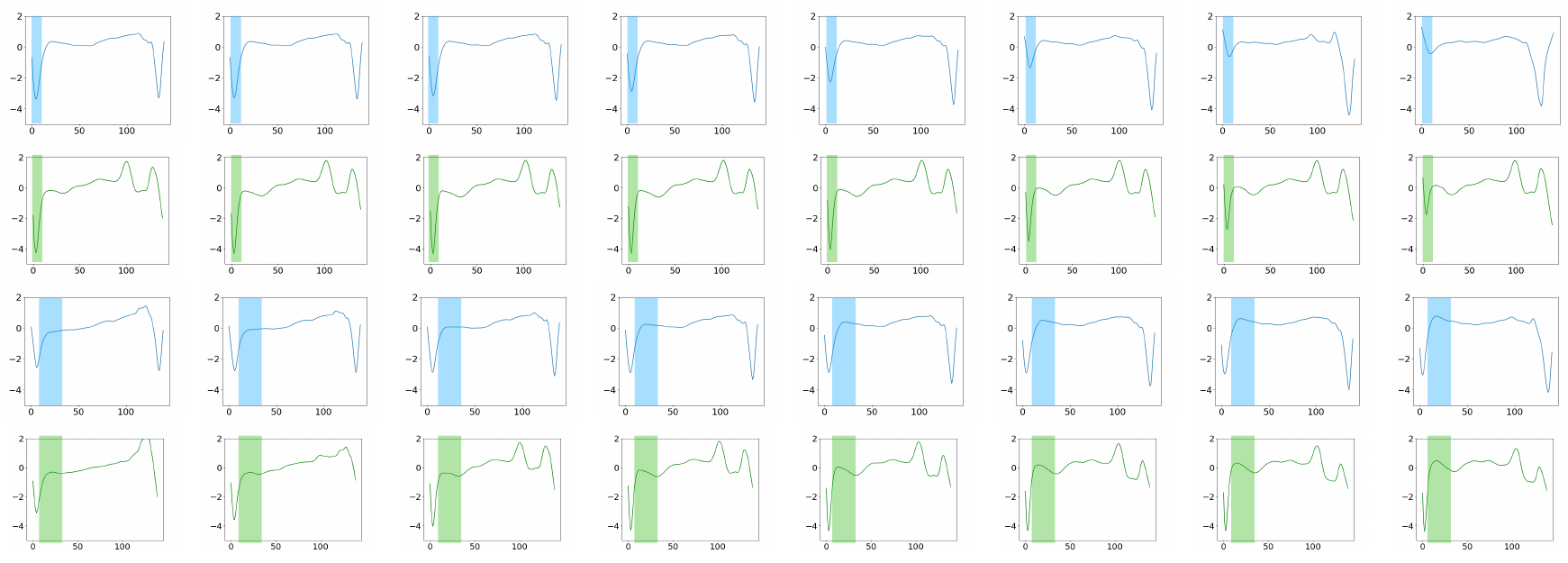}
        \vspace{-18pt}
    \caption{Latent traversal plots from \model on ECG. All figures of latent codes traversal each block corresponds to the traversal of a single latent variable while keeping others fixed to either their inferred. Each row represents a different seed image used to infer the latent values with traversal over the [-4, 4] range.  Blue and green denote two time-series with different sequential patterns. The first two rows denote the decline degree at the first turning point, transition from rigid to a more mild manner;  The last two rows denote the rising trend at the second turning point, transition from inconspicuous to obvious.}
    \label{fig:traversal}
    \vspace{-5pt}
\end{figure*}

\vspace{-5pt}
\subsubsection{Ablation Studies} 
We study whether \model can decompose the representations into domain-dependent and class-dependent components with a series of ablation studies.
We compare the discriminability of the disentangled features, including domain-dependent ($\mathbf{z}_d$) and class-dependent ($\mathbf{z}_y$) components (see Section~2.4.2), sampled from both source and target domains. The ablations include (i) using domain-dependent features $\mathbf{z}_d$ and class-dependent features $\mathbf{z}_y$, and (ii) domain-invariant and class-invariant features (shown as /r), respectively. 
The comparison between ~\model and the ablations (/r) is shown in Table~\ref{tab:ablation}. We observe that ~\model significantly outperforms the ablations over all datasets. Disentangled task-dependent group segments consistently help to improve the performance. Conversely, the class-dependent features are invariant  to the change of domains, and the domain-dependent features are invariant to the change of classes. 
 \model could preserve the discriminability of the disentangled features corresponding to the specified task, and simultaneously make disentangled features less discriminative for the irrelevant task. It indicates that these disentangled group segments do not contain any useful semantic concepts for other irrelevant tasks.

\subsection{Individual Latent Factor Disentanglement}

To answer the research question \textbf{Q2}, we provide latent traversals as qualitative results to validate that \model tends to consistently discover more informative latent factors and provide more meaningful disentangled representations of time-series (see Section~2.3).

\vspace{-3pt}
\subsubsection{Qualitative Comparison of Latent Codes} 
We train \model on ECG data to evaluate disentanglement performance for individual latent factors. We use
the same traversal way to show the disentanglement quality. Fig.~\ref{fig:latent} provides a qualitative comparison of the disentanglement performance of \model and LSTM-VAE. We edit the time-series by altering the latent codes in the $\mathcal{Z}$ space. Here, the dimension of the representation is set to be $12$. This setting helps reduce the impact of differences in complexity 
by model frameworks. However, for a better comparison, we only select eight dimensions that change more regularly. The sequences visualized in panels are generated from ${Z} \sim q\left({Z} \mid \boldsymbol{x}_{1:T}\right)$. Hence, the dynamics are imposed by the encoder, but the identity is sampled from the prior.


Fig.~\ref{fig:latent} shows traversals in latent variables that depict an interpretable property in generating time-series. Often it could generate more semantic convincing time-series than LSTM-VAE. Comparing the visualization from panel (a) and panel (b), \model could generate a time-series in a more diverse way. It can be seen that sampling from an entangled representation results in LSTM-VAE (panel (a)) only reflects small differences according to the traversal perturbations. 
One possible reason is that the LSTM-VAE is dominated by the reconstruction term during the training phase. The slight changes only correspond to the reconstruction distortion due to the latent codes and observations are independent. Comparing with LSTM-VAE, some of the \model latent codes tend to learn a smooth continuous transformation 
over a wider range of factor values as vibrations. A clear transition process can be observed from the manipulation results with respect to the $\mathcal{Z}$ space. As the value of individual latent factor increases, the semantic of the latent factor changes across different sequential patterns.  Other latent codes are robust with the vibrations, as it does not play any role in the generation process. Single latent units are sensitive to changes in single generative factor, while being relatively invariant to changes in other factors. One possible reason is that the representation of the time-series could be effectively expressed with a few latent codes in the $\mathcal{Z}$. Individual latent factor disentanglement process may help us to recognize the useful latent codes, and  discard the redundant parts. All these results demonstrate that \model is able to disentangle useful knowledge from sequential data, which is more informative as interpretable factors in the latent space.

\vspace{-5pt}
\subsubsection{Traversal Plots to Discover Semantics} There is currently no general method for quantifying the degree of learned disentanglement or  optimize the hyperparameters (unless there are concept ground-truth factors $v$ available, then the mutual information gap (MIG)~\cite{chen2018isolating,shao2020controlvae} could be used to determine if there exists a deterministic, invertible relationship between $z$ and $v$). Therefore, there is no way to quantitatively compare the degree of disentanglement achieved by different models or when optimizing the hyperparameters of a single model. Fig.~\ref{fig:traversal} plots the manipulation results of the latent traversal results from~\modelnospace. Each block of the figure corresponds  to the traversal of a single latent variable while keeping others fixed. Each row represents a different seed image used to infer the latent values with traversal over the [-4, 4] range. The results show that our manipulation approach performs well on all attributes in both positive and negative directions. We observe that moving the latent codes can produce continuous change, with the sequential patterns orthogonal to the others. According to the editing process, the first and the third rows (sampled from two different time-series of ECG) denote the decline degree at the first turning point, transition from rigid to a more mild manner;  the second and the last rows (with the same sampling strategy) denote the rising trend at the second turning point, transition from inconspicuous to obvious. It demonstrates that \model  discovered latent factors in an unsupervised manner that encode sequential trend and depict an interpretable property in the generation. 
These observations provide strong evidence that~\model does not produce time-series randomly, but learns some interpretable semantics in the latent space.

\begin{figure}[t]     
\vspace{-11pt}
\begin{subfigure}[b]{0.45\textwidth}
 \includegraphics[width=\linewidth]{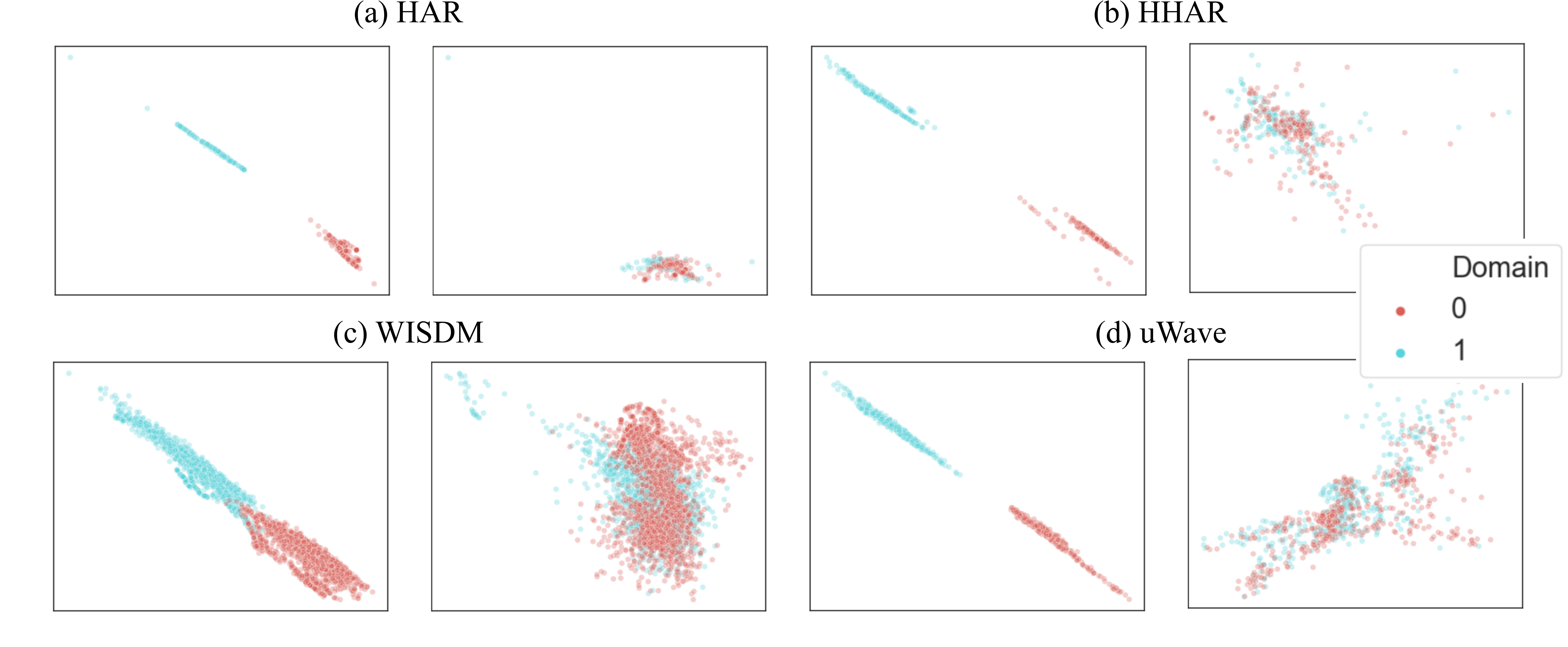}

             \caption{T-SNE visualizations of the \model activations on the distribution of domain-dependent representation (left) 
             and domain-invariant representations(right). 
             Blue points correspond to the source domain examples, while red ones correspond to the target domain ones. }
        \end{subfigure}%
        \newline
        \begin{subfigure}[b]{0.45\textwidth} \includegraphics[width=\linewidth]{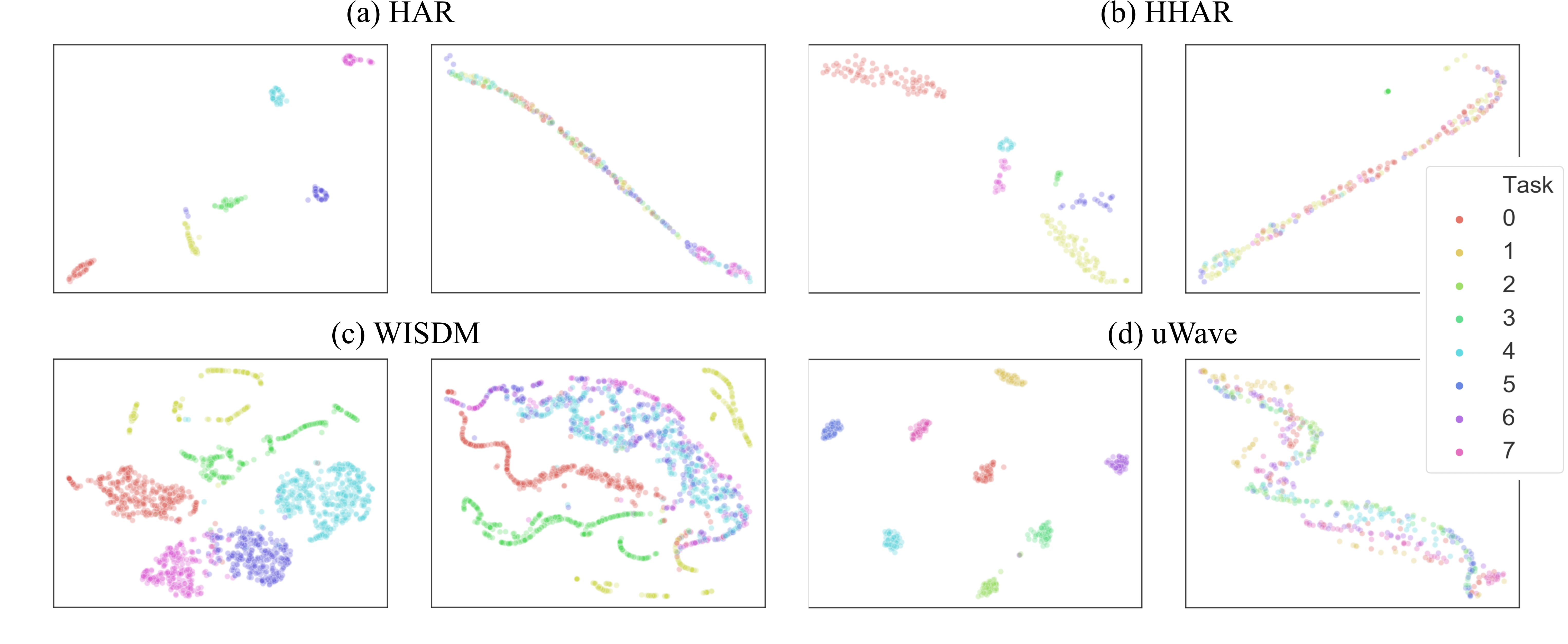}

                \caption{T-SNE visualizations of \model activations on the distribution of class-dependent representation (left)  
                and class-invariant representations (right).
                Each color denotes  one specific class. }
        \end{subfigure}%
              \vspace{-8pt}
        \caption{The effect of (a) domain-dependent and -invariant  (b) class-dependent and -invariant disentanglement on the distribution of the extracted features. In all cases, the adaptation in \model makes the dependent/invariant features from different sources more/less distinguishable, respectively. }
 \label{fig:domainviz}
\vspace{-5pt}
\end{figure}


\subsection{Latent Group Segment Disentanglement}
To answer the research question \textbf{Q3}, we visualize the disentangled segments over the representation space (see Section~2.4.1). Fig.~\ref{fig:domainviz}(a) shows the effect of domain-dependent and domain-invariant disentanglement on the distribution of the extracted features. We observe that, for all the datasets, the adaptation in \model makes the disentangled domain(class)-dependent features more distinguishable, but the domain(class)-invariant features indistinguishable. The results validate the ~\model can learn decomposed segments that contain independent semantic information. Furthermore, we can observe an apparent clustering effect (the different colors denote different categories). A widely-accepted assumption~\cite{ben2014domain} indicates that observation distribution contains separated data clusters and data samples in the same cluster share the same class label in domain adaptations. These results validate the discriminative ability of the disentanglement, since ~\model is capable of yielding strong clustering in the target domain. And it almost matches the prior perfectly, as the semantically similar observations are mapped closer, and create clusters. This phenomenon gives us another insight as the disentangled group segments could enhance the interpretability.

\vspace{-3pt}

\section{Conclusion}
In this paper, we investigated a novel and challenging problem of learning disentangled time-series representations. 
\model introduces a multi-level disentanglement strategy, covering both individual latent factor and group semantic segments, to generate hierarchical semantic concepts as the interpretable and disentangled representation. It alleviates the KL vanishing problem via introducing a mutual information maximization term, while preserving a heavier penalty on the total correlation and the dimension-wise KL to keep the disentangle property via balancing the preference between correct inference and fitting data distribution. 
The experimental results demonstrated the effectiveness of \model in learning interpretable semantic concepts with disentanglement.

\vspace{-5pt}
\bibliographystyle{ieeetr}
\bibliography{sample-base}

\begin{thebibliography}{10}

\bibitem{shen2020interfacegan}
Y.~Shen, C.~Yang, X.~Tang, and B.~Zhou, ``Interfacegan: Interpreting the
  disentangled face representation learned by gans,'' {\em TPAMI}, 2020.

\bibitem{blackburn1960electrocardiogram}
H.~Blackburn and et~al, ``The electrocardiogram in population studies: a
  classification system,''

\bibitem{lai2020tods}
K.-H. Lai, D.~Zha, G.~Wang, J.~Xu, Y.~Zhao, D.~Kumar, Y.~Chen, P.~Zumkhawaka,
  M.~Wan, D.~Martinez, {\em et~al.}, ``Tods: An automated time series outlier
  detection system,'' {\em arXiv preprint arXiv:2009.09822}, 2020.

\bibitem{zha2020meta}
D.~Zha, K.-H. Lai, M.~Wan, and X.~Hu, ``Meta-aad: Active anomaly detection with
  deep reinforcement learning,'' {\em arXiv preprint arXiv:2009.07415}, 2020.

\bibitem{liu2019single}
N.~Liu, Q.~Tan, Y.~Li, H.~Yang, J.~Zhou, and X.~Hu, ``Is a single vector
  enough? exploring node polysemy for network embedding,'' in {\em Proceedings
  of the 25th ACM SIGKDD International Conference on Knowledge Discovery \&
  Data Mining}, 2019.

\bibitem{huang2019graph}
X.~Huang, Q.~Song, Y.~Li, and X.~Hu, ``Graph recurrent networks with attributed
  random walks,'' in {\em Proceedings of the 25th ACM SIGKDD International
  Conference on Knowledge Discovery \& Data Mining}, 2019.

\bibitem{du2019towards}
M.~Du, S.~Pentyala, Y.~Li, and X.~Hu, ``Towards generalizable forgery detection
  with locality-aware autoencoder,'' {\em arXiv preprint arXiv:1909.05999},
  2019.

\bibitem{lai2020dual}
K.-H. Lai, D.~Zha, Y.~Li, and X.~Hu, ``Dual policy distillation,'' {\em arXiv
  preprint arXiv:2006.04061}, 2020.

\bibitem{zhou2020towards}
K.~Zhou, X.~Huang, Y.~Li, D.~Zha, R.~Chen, and X.~Hu, ``Towards deeper graph
  neural networks with differentiable group normalization,'' {\em arXiv
  preprint arXiv:2006.06972}, 2020.

\bibitem{li2019specae}
Y.~Li, X.~Huang, J.~Li, M.~Du, and N.~Zou, ``Specae: Spectral autoencoder for
  anomaly detection in attributed networks,'' in {\em Proceedings of the 28th
  ACM International Conference on Information and Knowledge Management}, 2019.

\bibitem{li2020pyodds}
Y.~Li, D.~Zha, P.~Venugopal, N.~Zou, and X.~Hu, ``Pyodds: An end-to-end outlier
  detection system with automated machine learning,'' in {\em Companion
  Proceedings of the Web Conference 2020}, 2020.

\bibitem{li2019deep}
Y.~Li, N.~Liu, J.~Li, M.~Du, and X.~Hu, ``Deep structured cross-modal anomaly
  detection,'' in {\em 2019 International Joint Conference on Neural Networks
  (IJCNN)}, 2019.

\bibitem{li2020autood}
Y.~Li, Z.~Chen, D.~Zha, K.~Zhou, H.~Jin, H.~Chen, and X.~Hu, ``Autood:
  Automated outlier detection via curiosity-guided search and self-imitation
  learning,'' {\em arXiv preprint arXiv:2006.11321}, 2020.

\bibitem{semeniuta2017hybrid}
S.~Semeniuta and et~al, ``A hybrid convolutional variational autoencoder for
  text generation,'' {\em arXiv preprint arXiv:1702.02390}, 2017.

\bibitem{fortuin2018som}
V.~Fortuin and et~al, ``Som-vae: Interpretable discrete representation learning
  on time series,'' {\em arXiv preprint arXiv:1806.02199}, 2018.

\bibitem{fortuin2020gp}
V.~Fortuin, D.~Baranchuk, G.~R{\"a}tsch, and S.~Mandt, ``Gp-vae: Deep
  probabilistic time series imputation,'' in {\em AISTAT}, 2020.

\bibitem{yang2019xdeep}
F.~Yang, Z.~Zhang, H.~Wang, Y.~Li, and X.~Hu, ``Xdeep: An interpretation tool
  for deep neural networks,'' {\em arXiv preprint arXiv:1911.01005}, 2019.

\bibitem{guo2020interpretable}
X.~Guo, L.~Zhao, Z.~Qin, L.~Wu, A.~Shehu, and Y.~Ye, ``Interpretable deep graph
  generation with node-edge co-disentanglement,'' in {\em KDD}, 2020.

\bibitem{shao2020controlvae}
H.~Shao, S.~Yao, D.~Sun, A.~Zhang, S.~Liu, D.~Liu, J.~Wang, and T.~Abdelzaher,
  ``Controlvae: Controllable variational autoencoder,'' in {\em ICML}, 2020.

\bibitem{chen2018isolating}
R.~T. Chen, X.~Li, R.~Grosse, and D.~Duvenaud, ``Isolating sources of
  disentanglement in variational autoencoders,'' {\em arXiv preprint
  arXiv:1802.04942}, 2018.

\bibitem{wang2019riemannian}
P.~Z. Wang and W.~Y. Wang, ``Riemannian normalizing flow on variational
  wasserstein autoencoder for text modeling,'' {\em arXiv preprint
  arXiv:1904.02399}, 2019.

\bibitem{chen2016variational}
X.~Chen, D.~P. Kingma, T.~Salimans, Y.~Duan, P.~Dhariwal, J.~Schulman,
  I.~Sutskever, and P.~Abbeel, ``Variational lossy autoencoder,'' {\em
  arXiv:1611.02731}, 2016.

\bibitem{chung2015recurrent}
J.~Chung, K.~Kastner, L.~Dinh, K.~Goel, A.~Courville, and Y.~Bengio, ``A
  recurrent latent variable model for sequential data,'' {\em arXiv preprint
  arXiv:1506.02216}, 2015.

\bibitem{kingma2013auto}
D.~P. Kingma and M.~Welling, ``Auto-encoding variational bayes,'' {\em arXiv
  preprint arXiv:1312.6114}, 2013.

\bibitem{burgess2018understanding}
C.~P. Burgess, I.~Higgins, A.~Pal, L.~Matthey, N.~Watters, {\em et~al.},
  ``Understanding disentangling in beta-vae,'' {\em arXiv preprint
  arXiv:1804.03599}, 2018.

\bibitem{watanabe1960information}
S.~Watanabe, ``Information theoretical analysis of multivariate correlation,''
  {\em IBM Journal of research and development}, 1960.

\bibitem{zhao2019infovae}
S.~Zhao, J.~Song, and S.~Ermon, ``Infovae: Balancing learning and inference in
  variational autoencoders,'' in {\em AAAI}, 2019.

\bibitem{hinton1994autoencoders}
G.~E. Hinton and R.~S. Zemel, ``Autoencoders, minimum description length, and
  helmholtz free energy,'' {\em NeurIPS}, 1994.

\bibitem{bai2018empirical}
S.~Bai {\em et~al.}, ``An empirical evaluation of generic convolutional and
  recurrent networks for sequence modeling,'' {\em arXiv preprint
  arXiv:1803.01271}, 2018.

\bibitem{ganin2015unsupervised}
Y.~Ganin and V.~Lempitsky, ``Unsupervised domain adaptation by
  backpropagation,'' in {\em ICML}, 2015.

\bibitem{cai2019learning}
R.~Cai, Z.~Li, P.~Wei, J.~Qiao, K.~Zhang, and Z.~Hao, ``Learning disentangled
  semantic representation for domain adaptation,'' in {\em IJCAI}, 2019.

\bibitem{locatello2019challenging}
F.~Locatello and et~al, ``Challenging common assumptions in the unsupervised
  learning of disentangled representations,'' in {\em ICML}, 2019.

\bibitem{tishby2015deep}
N.~Tishby and N.~Zaslavsky, ``Deep learning and the information bottleneck
  principle,'' in {\em 2015 IEEE Information Theory Workshop (ITW)}, 2015.

\bibitem{tzeng2014deep}
E.~Tzeng and et~al, ``Deep domain confusion: Maximizing for domain
  invariance,'' {\em arXiv preprint arXiv:1412.3474}, 2014.

\bibitem{wilson2020multi}
G.~Wilson, J.~R. Doppa, and D.~J. Cook, ``Multi-source deep domain adaptation
  with weak supervision for time-series sensor data,'' in {\em KDD}, 2020.

\bibitem{anguita2013public}
D.~Anguita, A.~Ghio, L.~Oneto, X.~Parra, and J.~L. Reyes-Ortiz, ``A public
  domain dataset for human activity recognition using smartphones.,'' in {\em
  Esann}, 2013.

\bibitem{stisen2015smart}
A.~Stisen and et~al, ``Smart devices are different: Assessing and
  mitigatingmobile sensing heterogeneities for activity recognition,'' in {\em
  SenSys}, 2015.

\bibitem{kwapisz2011activity}
J.~R. Kwapisz, G.~M. Weiss, and S.~A. Moore, ``Activity recognition using cell
  phone accelerometers,'' {\em ACM SigKDD Explorations Newsletter}, 2011.

\bibitem{liu2009uwave}
J.~Liu and et~al, ``uwave: Accelerometer-based personalized gesture recognition
  and its applications,'' {\em Pervasive and Mobile Computing}, 2009.

\bibitem{ajakan2014domain}
H.~Ajakan, P.~Germain, H.~Larochelle, F.~Laviolette, and M.~Marchand,
  ``Domain-adversarial neural networks,'' {\em arXiv preprint arXiv:1412.4446},
  2014.

\bibitem{purushotham2017variational}
S.~Purushotham, W.~Carvalho, T.~Nilanon, and Y.~Liu, ``Variational recurrent
  adversarial deep domain adaptation.,'' in {\em ICLR}, 2017.

\bibitem{ben2014domain}
S.~Ben-David and R.~Urner, ``Domain adaptation--can quantity compensate for
  quality?,'' {\em Annals of Mathematics and Artificial Intelligence}, 2014.

\end{thebibliography}

\end{document}